\documentclass[letterpaper]{article} 
\usepackage{aaai25}  
\usepackage{times}  
\usepackage{helvet}  
\usepackage{courier}  
\usepackage[hyphens]{url}  
\usepackage{graphicx} 
\usepackage{natbib}  
\usepackage{caption} 
\usepackage{algorithm}
\usepackage{algorithmic}
\usepackage{amsfonts}
\usepackage{amsmath}
\usepackage{multirow}
\usepackage{arydshln}
\usepackage{subfigure}
\usepackage{algorithm}
\usepackage{algorithmic}
\usepackage{xcolor}
\usepackage{newfloat}
\usepackage{listings}
\DeclareCaptionStyle{ruled}{labelfont=normalfont,labelsep=colon,strut=off} 
\floatstyle{ruled}
\newfloat{listing}{tb}{lst}{}
\floatname{listing}{Listing}
\title{ComprehendEdit: A Comprehensive Dataset and Evaluation Framework for Multimodal Knowledge Editing}
\author {
    Yaohui Ma\textsuperscript{\rm 1, \rm 3},
    Xiaopeng Hong\textsuperscript{\rm 1, \rm 2},
    Shizhou Zhang\textsuperscript{\rm 4},
    Huiyun Li\textsuperscript{\rm 3, \rm 5, \rm 6},\\
    Zhilin Zhu\textsuperscript{\rm 1, \rm 2},
    Wei Luo\textsuperscript{\rm 3},
    Zhiheng Ma\textsuperscript{\rm 3, \rm 5, \rm 6}\thanks{Corresponding author: Zhiheng Ma (zh.ma@siat.ac.cn)}
}
\affiliations {
    \textsuperscript{\rm 1}Faculty of Computing, Harbin Institute of Technology \quad
    \textsuperscript{\rm 2}Pengcheng Laboratory\\
    \textsuperscript{\rm 3}Shenzhen University of Advanced Technology\\
    \textsuperscript{\rm 4}School of Computer Science, Northwestern Polytechnical University\\

    \textsuperscript{\rm 5}Guangdong Provincial Key Laboratory of Computility Microelectronics\\

    \textsuperscript{\rm 6}Shenzhen Institutes of Advanced Technology, Chinese Academy of Sciences\\

    \{yhma, zhuzhilin\}@stu.hit.edu.cn, hongxiaopeng@hit.edu.cn, szzhang@nwpu.edu.cn, \\
    \{hy.li, zh.ma\}@siat.ac.cn, luow0014@e.ntu.edu.sg
}

\usepackage{bibentry}

\begin{document}
\maketitle

\begin{abstract}
Large multimodal language models (MLLMs) have revolutionized natural language processing and visual understanding, but often contain outdated or inaccurate information. Current multimodal knowledge editing evaluations are limited in scope and potentially biased, focusing on narrow tasks and failing to assess the impact on in-domain samples. To address these issues, we introduce ComprehendEdit, a comprehensive benchmark comprising eight diverse tasks from multiple datasets. We propose two novel metrics: Knowledge Generalization Index (KGI) and Knowledge Preservation Index (KPI), which evaluate editing effects on in-domain samples without relying on AI-synthetic samples. Based on insights from our framework, we establish Hierarchical In-Context Editing (HICE), a baseline method employing a two-stage approach that balances performance across all metrics. This study provides a more comprehensive evaluation framework for multimodal knowledge editing, reveals unique challenges in this field, and offers a baseline method demonstrating improved performance. Our work opens new perspectives for future research and provides a foundation for developing more robust and effective editing techniques for MLLMs. The ComprehendEdit benchmark and implementation code are available at \url{https://github.com/yaohui120/ComprehendEdit}.
\end{abstract}

\section{Introduction}
The advent of large language models (LLMs) has transformed natural language processing~\cite{zhao2023survey}, while exposing limitations in maintaining up-to-date information and rectifying inaccuracies~\cite{dhingra2022time, elazar2021measuring, cao2021knowledgeable}. To address these challenges, knowledge editing methods~\cite{zheng2023can, sun2024outdated, chen2024lifelong, de2021editing, meng2022locating, deng2024unke, hu2024wilke, mitchell2022memory, mitchell2021fast, huang2023transformer} enable updating outdated or incorrect knowledge within LLMs without complete retraining. These methods primarily focus on achieving reliability (successfully editing specified problems), generality (appropriately adjusting answers to similar questions), and locality (maintaining consistent responses to unrelated questions).

\begin{figure}[t]%
\centering
\includegraphics[width=0.47\textwidth]{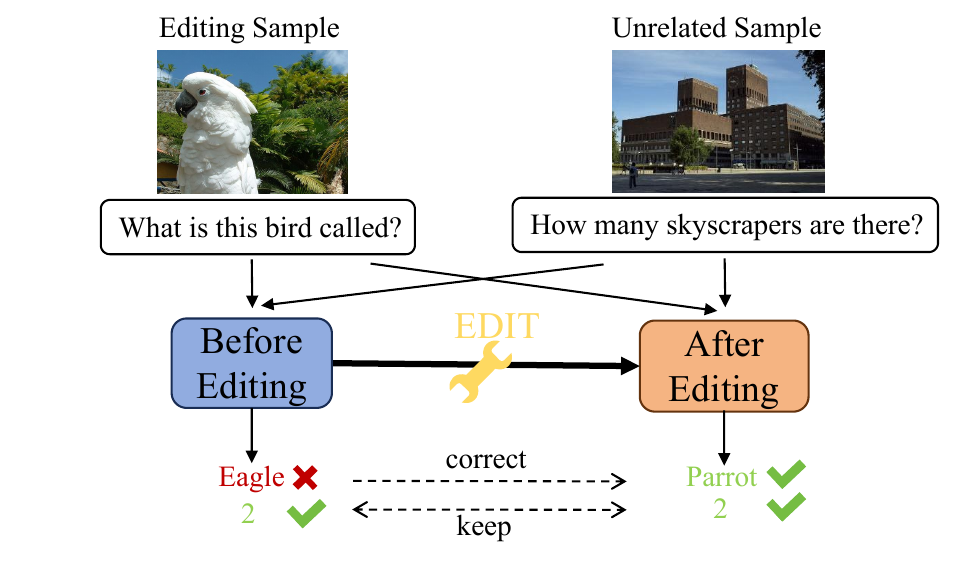}
\caption{Concept of Multimodal Knowledge Editing. The framework is to correct the wrong answer for the editing sample (``Eagle'' to ``Parrot'') while maintaining the output for unrelated samples (``2'' to ``2'').}
\label{fig:intro_multimodal_edit}
\end{figure}

As multimodal large language models (MLLMs) emerge, new challenges arise in knowledge editing. While MLLMs like BLIP-2 OPT~\cite{han2023divide}, MiniGPT-4~\cite{zhu2023minigpt}, Qwen-VL~\cite{bai2023qwen} and LLaVA-1.5~\cite{liu2024visual} excel at answering questions about images, they still exhibit errors and misunderstandings. These inaccuracies stem from both language and vision modules~\cite{liu2024survey, rawte2024visual, tong2024eyes, jiang2023clip}, necessitating multimodal-specific editing techniques.

Recent studies like \citeauthor{cheng2023edit} have established evaluation frameworks for multimodal knowledge editing through their MMEdit benchmark (including E-VQA and E-IC~\cite{cheng2023edit}), which builds upon VQAv2~\cite{goyal2017making} and COCO Caption~\cite{chen2015microsoft}. They assess methods on reliability in modifying target outputs, generality across rephrased questions~\cite{du2021glm} and generated images~\cite{rombach2022high}, and locality in preserving responses on out-of-domain datasets like NQ dataset~\cite{kwiatkowski2019natural} and OK-VQA~\cite{marino2019ok}. Initial results are promising - transferring language model editing techniques to multimodal contexts has proven effective, with methods like MEND~\cite{mitchell2021fast} achieving $98.51\%$ reliability and $96.65\%$ multimodal locality on E-VQA~\cite{cheng2023edit}.

\begin{figure}[t]
\centering
\includegraphics[width=0.49\textwidth]{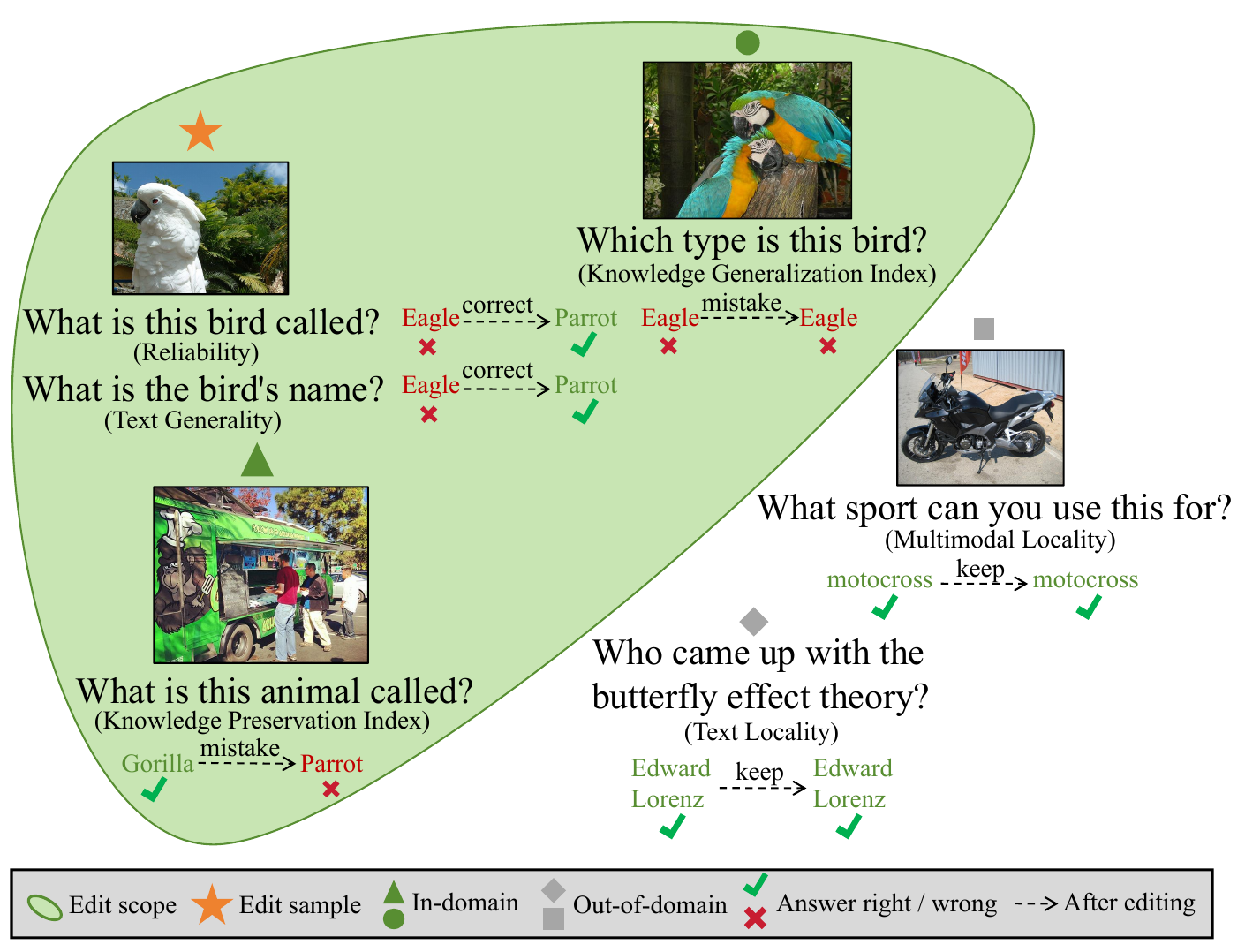}
\caption{Knowledge Distortion in Multimodal Knowledge Editing. It shows how the model maintains correct outputs for out-of-domain samples but struggles with in-domain samples, highlighting the challenge in preserving and generalizing knowledge.}
\label{fig:intro_distortion}
\end{figure}

However, we argue that current multimodal knowledge editing evaluations are incomplete and potentially biased for the following reasons:

1) Limited Task Coverage: Existing assessments like E-VQA focus on narrow tasks, failing to capture broad MLLM capabilities such as spatial reasoning.

2) AI-synthetic Content Issues: Generating equivalent images introduces unpredictable content shifts~\cite{huang2024kebench}, while VQA questions offer limited rephrasing variations (e.g., ``Is there a tree in front of the building?'' vs ``What is the status of the tree in relation to the building?'').

3) Out-of-domain Only Evaluation: Current locality assessment uses only distant, unrelated samples, missing potential unintended changes to in-domain knowledge.

To address these limitations, we introduce ComprehendEdit with three key innovations:

1) Comprehensive Task Coverage: Eight diverse tasks derived from multiple datasets, ensuring broad evaluation of MLLM capabilities.

2) No Synthetic Content Dependency: Two novel metrics -- Knowledge Generalization Index (KGI) and Knowledge Preservation Index (KPI) that evaluate editing effects without relying on AI-synthetic content.

3) In-domain Assessment: These metrics specifically measure how editing affects similar knowledge within the same domain, providing crucial insights previously overlooked.

Within this evaluation framework, we thoroughly assessed existing multimodal knowledge editing methods. Our findings reveal that current approaches struggle to perform optimally across all metrics, indicating significant room for improvement. Many methods that excelled in previous evaluations performed poorly on the new metrics, demonstrating both the bias in earlier assessments and the considerable potential for advancement in multimodal knowledge editing techniques.

Based on the issues revealed by our evaluation framework, we establish a baseline method, Hierarchical In-Context Editing (HICE), and conduct comprehensive ablation studies to investigate various trade-offs in multimodal knowledge editing. HICE achieves comparable accuracy to previous state-of-the-art methods on existing metrics, while demonstrating superior and more balanced performance on the newly introduced metrics.

In summary, this study advances multimodal knowledge editing research by revealing unique challenges distinct from language-only editing, particularly in preserving and generalizing knowledge within the editing domain. Through our comprehensive framework ComprehendEdit, we establish new metrics for evaluating knowledge effects on editing-related samples, while providing a strong baseline method for systematic comparison. This work not only exposes previously overlooked deficiencies in current approaches but also establishes a foundation for developing more effective multimodal knowledge editing techniques.

\section{Related works}
\subsection{Knowledge Editing}
Recent studies on knowledge editing can be classified into three categories: locate and update knowledge neurons, meta learning methods and memory based methods.

\textbf{Locate and update methods} focus on identifying and modifying specific neurons within a model. ROME~\cite{meng2022locating} applies the interventions on activation to determine which neurons have the strongest effect on the prediction. KN~\cite{dai2021knowledge} uses the integrated gradients method to calculate neuron contributions, thereby identifying knowledge neurons. T-patcher~\cite{huang2023transformer} locates and inserts trainable neurons into specific layers to alter the model's output. Additionally, UnKE~\cite{deng2024unke} and WilKE~\cite{hu2024wilke} are not restricted to particular MLP layers or knowledge neurons. They search for parameters to edit across a broader range of locations.

Although these methods update only a few of neurons, they require substantial computation to identify the location of updated neurons, which increases the training cost. Additionally, their application in black box models is limited.

\textbf{Meta learning methods} employ auxiliary models to guide parameter updates. MEND~\cite{mitchell2021fast} and KE~\cite{de2021editing} both train additional modules to adjust gradients, ensuring that the optimized model minimally impacts predictions for unrelated inputs. SLAG~\cite{hase2023methods} uses LSTM and MLPs to learn a set of weights for gradient modification.

Compared to locate-and-update methods, meta learning methods demonstrate superior locality. However, the space and training time required for the additional networks are significant considerations.

\textbf{Memory based methods} store learnable parameters and editing samples within training set. SERAC~\cite{mitchell2022memory} stores an editing sample and trains a counterfactual model to obtain the expected output. IKE~\cite{zheng2023can} first employ in-context learning~\cite{brown2020language} for editing knowledge in language models. They construct demonstrations of each training sample, and select appropriate demonstrations as context to modify the model's output. DISCO~\cite{sun2024outdated} similarly use in-context learning to enhance the edited model's ability to utilize the edited knowledge for reasoning. Building upon these approaches, HICE introduces a two-stage process: it first classifies samples as in-domain or out-of-domain before applying in-context learning. This classification step prevents the application of in-context learning to out-of-domain samples, thus avoiding potential interference from irrelevant demonstrations. 

\subsection{Multimodal knowledge editing}

The advancement of Multimodal Large Language Models (MLLMs) demands new approaches to knowledge editing. \citeauthor{cheng2023edit} first introduced multimodal knowledge editing and developed a benchmark named MMEdit. They also established novel evaluation metrics: reliability, generality, and locality. Similarly, KEBench~\cite{huang2024kebench} extends existing metrics and introduces a portability metric to assess the model's ability to effectively apply edited knowledge to related content.

\subsection{Benchmark for Multimodal Large Language Models}

Recently, datasets used to evaluate multimodal large language models typically encompass assessments of various abilities, such as perception (e.g., object existence, quantity, and attributes) and reasoning (e.g., common sense reasoning and numerical calculation). However, most of these datasets are not suitable for knowledge editing evaluation. They either contain too few samples to support trainable methods (e.g., POPE~\cite{li2023evaluating}, MME~\cite{fu2023mme}) or cannot be evaluated offline (e.g., VizWiz~\cite{gurari2018vizwiz}, MMBench~\cite{liu2023mmbench}, MM-vet~\cite{yu2023mm}).

Other datasets that can be used have limited types of model capability evaluations. GQA~\cite{hudson2019gqa} offers vast samples but they are mostly limited to object existence, object recognition, object attributes and scene information. Other datasets focus exclusively on evaluating a certain capability of the model. For instance, TextVQA~\cite{singh2019towards} focuses on assessing the model's ability to recognize text. TallyQA~\cite{acharya2019tallyqa} consists of object counting questions, with complex types that require content understanding. VSR~\cite{liu2023visual} emphasizes the spatial relationship between objects, encompassing dozens of relationships. MathVista~\cite{lu2023mathvista} collects various graphs and tables, all of which require numerical reasoning to answer correctly. In order to overcome these drawbacks and to evaluate multimodal knowledge editing methods more comprehensively, we construct a novel benchmark ComprehendEdit.

\section{Proposed Method}

\subsection{Dataset}

\citeauthor{cheng2023edit} was the first to propose the multimodal editing problem and developed two tasks E-VQA and E-IC. However, \citeauthor{huang2024kebench} identified content shifts in the generated images within these datasets, leading to inaccurate locality assessments. Additionally, due to the extensive capabilities of MLLMs, a single-source dataset is inadequate for evaluating knowledge editing methods comprehensively. Existing datasets inadequately address diverse editing challenges due to their limited variety in question types and sample diversity, as shown in appendix. Commonly used MLLMs evaluation datasets, such as VizWiz~\cite{gurari2018vizwiz}, MMBench~\cite{liu2023mmbench}, MME~\cite{fu2023mme}, MM-vet~\cite{yu2023mm}, POPE~\cite{li2023evaluating}, are unsuitable for evaluating knowledge editing methods due to insufficient training samples or inability to support offline evaluation. To overcome these limitations, we propose a new benchmark ComprehendEdit, which comprises 8 tasks derived from diverse datasets. The details of the dataset are shown in Table~\ref{tab:subtask_info}.

\begin{table}[h] 
\begin{center}
\caption{Task Distribution in ComprehendEdit. It details the number of training and testing samples for each task.}
\label{tab:subtask_info}
\resizebox{.95\columnwidth}{!}{
\begin{tabular}{cccc}
\hline
Task   & Training set & Testing set & Source  \\
\hline
Object Existence & 1471 & 491 & GQA \\
Object Recognition & 2227 & 735 & GQA \\
Object Attributes & 2282 & 705 & GQA \\
Object Counting & 1506 & 503 & TallyQA \\
Scene Information & 2067 & 787 & GQA \\
Spatial Relationship & 1709 & 530 & VSR \\
Text Recognition & 1554 & 519 & TextVQA \\
Numerical Inference & 634 & 212 & MathVista \\
\hline
Total & 13450 & 4482 & \\
\hline
\end{tabular}}
\end{center}
\end{table}

The ComprehendEdit benchmark encompasses eight diverse tasks.
The ratio of training data to test data in each task is approximately 3:1, with a total of 17,932 samples. Examples from the dataset and detailed construction of each task are provided in the appendix.

For measuring text generality, we use a pre-trained model (such as ChatGLM~\cite{du2021glm}) to generate equivalent inputs (rephrased questions). Additionally, we also utilize samples from the NQ dataset~\cite{kwiatkowski2019natural} and OK-VQA dataset~\cite{marino2019ok} to measure text locality (T-L) and multimodal locality (M-L), respectively, following previous benchmarks~\cite{cheng2023edit}.

\subsection{Task Formulation}
The goal of multimodal knowledge editing is to adjust the output of a multimodal language model for a specific sample. To formalize this goal, we consider that an editing dataset $\mathcal{D}_e$ containing $N$ samples. Each editing sample $s$ comprises an image content $i_e$, a text question $x_e$, and a ground-truth $y_e$, represented as $(i_e, x_e, y_e)$. Additionally, for each $s$, a rephrased question, a locality sample and a multimodal locality sample are also provided. The parameters of the model $f$ before and after editing are denoted as $\theta_o, \theta_e$, respectively.

\subsection{Conventional Evaluation Metrics}

\textbf{Reliability}. Reliability ($\mathcal{M}_{rel}$) measures how effectively a model's knowledge can be edited. Given an editing dataset $\mathcal{D}_e$, for each sample $(i_e,x_e,y_e)$, the goal is to modify the model $f$ with parameters $\theta_o$ such that its output changes from the original incorrect prediction $y_o = f(i_e,x_e;\theta_o)$ to the desired correct answer $y_e$ after editing (with parameters $\theta_e$). Reliability is formally defined as:

\begin{equation}
 \mathcal{M}_{rel} = \underset{(i_e,x_e,y_e) \in \mathcal{D}_e}{\mathbb{E}} \mathbb{I}(f(i_e,x_e;\theta_e)=y_e),
 \label{eq:m_rel}
\end{equation}

\noindent where $\mathbb{I}(\cdot)$ is an indicator function that returns 1 if the edited model's output matches the target answer and 0 otherwise.

\textbf{Generality}. Generality assesses whether the editing effects can transfer to semantically equivalent inputs. Beyond the original editing sample $(i_e, x_e)$, the model should maintain correct behavior on variations of both the question and image while preserving the same meaning.

Following \citeauthor{cheng2023edit}, we generate equivalent variations using pre-trained models: rephrased questions $x_r$ using LLMs~\cite{du2021glm} (e.g., ``What color is the floor?'' → ``What color is the ground?''), and alternative images $i_r$ using diffusion models~\cite{rombach2022high}. Let $\mathcal{N}(x_e)$ and $\mathcal{N}(i_e)$ denote the sets of generated questions and images respectively. We evaluate text generality (T-G) and multimodal generality (M-G) as:

\begin{equation}
 \mathcal{M}_{general}^{txt} = \underset{\underset{x \in \mathcal{N}(x_e)}{(i_e,x_e,y_e) \in \mathcal{D}_e}}{\mathbb{E}} \mathbb{I}(f(i_e,x;\theta_e)=y_e),
 \label{eq:m_txt_general}
\end{equation}

\begin{equation}
 \mathcal{M}_{general}^{img} = \underset{\underset{i \in \mathcal{N}(i_e)}{(i,x_e,y_e) \in \mathcal{D}_e}}{\mathbb{E}} \mathbb{I}(f(i,x_e;\theta_e)=y_e).
 \label{eq:m_img_general}
\end{equation}

\noindent where $\mathcal{M}_{general}^{txt}$ measures performance on rephrased questions and $\mathcal{M}_{general}^{img}$ evaluates performance on generated images.

\textbf{Locality}. Locality measures whether knowledge editing preserves the model's behavior on unrelated inputs. Following \citeauthor{cheng2023edit}, we evaluate locality using external datasets: NQ dataset~\cite{kwiatkowski2019natural} for text questions and OK-VQA dataset~\cite{marino2019ok} for image-text questions. Text locality (T-L) and multimodal locality (M-L) are defined as:

\begin{equation}
 \mathcal{M}_{loc}^{txt} = \underset{(x,y) \in \mathcal{D}_{loc}}{\mathbb{E}} \mathbb{I}(f(x;\theta_e)=f(x;\theta_o)),
 \label{eq:m_txt_loc}
\end{equation}

\begin{equation}
 \mathcal{M}_{loc}^{img} = \underset{(i, x ,y) \in \mathcal{D}_{loc \text{-} v}}{\mathbb{E}} \mathbb{I}(f(i, x;\theta_e)=f(i, x;\theta_o)),
 \label{eq:m_multi_loc}
\end{equation}

\noindent where $\mathcal{D}_{loc}$ and $\mathcal{D}_{loc \text{-} v}$ are datasets containing samples significantly different from the edited samples. Note that locality measures output consistency rather than correctness - the original outputs $f(x;\theta_o)$ or $f(i, x;\theta_o)$ may be incorrect.

\subsection{Proposed Evaluation Metrics}
While conventional metrics focus on rephrased questions and out-of-domain samples, they overlook crucial aspects of knowledge editing within the same domain. Moreover, they rely on synthetic data that can introduce measurement inaccuracies through content shifts and semantic mismatches. To address these limitations, we propose two complementary metrics that directly evaluate editing effects on original in-domain samples.

Given an editing sample $s$, let $\mathcal{D}(s)$ denote the set of samples from the same source dataset as $s$. We split $\mathcal{D}(s)$ into complementary subsets such that $\mathcal{D}(s) = \mathcal{D}_{KGI}(s) \cup \mathcal{D}_{KPI}(s)$, where $\mathcal{D}_{KGI}(s)$ contains samples that the original model answered incorrectly (exclude $s$), and $\mathcal{D}_{KPI}(s)$ contains samples that the original model answered correctly.

\textbf{Knowledge Generalization Index (KGI)} measures how well the editing improves model performance on previously misclassified in-domain samples. For instance, after correcting the model to identify a specific parrot instead of ``eagle'', KGI evaluates whether this correction generalizes to other misclassified images. Unlike traditional generalization metrics that rely on synthetic data~\cite{huang2024kebench}, KGI uses real samples to avoid measurement artifacts:

\begin{equation}
  \begin{aligned}
 \mathcal{M}_{KGI} =\underset{s \in \mathcal{D}_{e}}{\mathbb{E}} \underset{s' \in \mathcal{D}_{KGI}(s)}{\mathbb{E}} \mathbb{I}(f(i', x';\theta_e)=y'),
  \end{aligned}
 \label{eq:KGI}
\end{equation}

\textbf{Knowledge Preservation Index (KPI)} assesses whether editing preserves the model's correct behavior on in-domain samples. It quantifies potential negative impacts where editing might disrupt previously correct predictions, such as changing a correct gorilla identification after editing bird-related knowledge. KPI is defined as:

\begin{equation}
  \begin{aligned}
 \mathcal{M}_{KPI} =\underset{s \in \mathcal{D}_{e}}{\mathbb{E}} \underset{s' \in \mathcal{D}_{KPI}(s)}{\mathbb{E}} \mathbb{I}(f(i', x';\theta_e)=y'),
  \end{aligned}
 \label{eq:kpi}
\end{equation}

\noindent where for both metrics, $s'=(i', x', y')$ represents an in-domain sample.

\textbf{Similarity-based Sampling}. While KGI and KPI provide comprehensive evaluation metrics, testing all in-domain samples after each editing operation incurs substantial computational costs. To address this efficiency challenge while maintaining metric effectiveness, we propose a similarity-based sampling strategy.

For each editing sample $s$, we select the $k$ most similar and $k$ most dissimilar samples from $\mathcal{D}_{KGI}(s)$ and $\mathcal{D}_{KPI}(s)$ based on either image or text similarity scores. This dual-ended sampling approach captures both the local and global effects of knowledge editing. Specifically, we compute: 1) Image-based metrics (I-KGI, I-KPI): using visual feature similarity between images; 2)Text-based metrics (T-KGI, T-KPI): using semantic similarity between questions.

This sampling strategy not only reduces computational overhead but also enables fine-grained analysis of how editing effects propagate differently through visual and linguistic domains. The high-similarity samples reveal local editing impacts, while low-similarity samples help assess potential far-reaching effects within the same domain.

\subsection{Hierarchical In-Context Editing}
Pre-trained models are sensitive to parameter changes, which can significantly impact their performance on in-domain samples. Two-stage methods~\cite{mitchell2022memory, hartvigsen2024aging, yu2024melo} address this by first determining whether an input requires a modified output and then generating the corresponding result. While IKE~\cite{zheng2023can} leverages contextual capabilities without modifying parameters, it can affect outputs on external data due to unrelated demonstrations. Inspired by IKE and two-stage approaches, we propose Hierarchical In-Context Editing (HICE). This method first determines if an input falls within the edited scope, then outputs either the original or updated prediction. This approach leverages contextual learning for in-domain data while preserving locality on external samples.

Recent studies suggest that features extracted by pre-trained models can be well adapted to classification tasks~\cite{panos2023first, mcdonnell2024ranpac}. Based on this, we use a pre-trained language model $h$ to extract text features for the first stage classification. To enhance classification accuracy, these features are projected into higher dimensions~\cite{mcdonnell2024ranpac}. Illustration of HICE is placed in the appendix.

Here we provide a detailed introduction to the method. For each sample $s$ in $\mathcal{D}_e$ along with its rephrased question, locality sample and multimodal locality sample, we follow IKE to structure these questions and answers separately into the template ``New Fact: $\{x\}$ $\{y\}$ $\backslash$n Prompt: \{$x$\} \{$y$\}'' to construct four demonstrations. Each demonstration is labeled with a one-hot vector $Y \in \{0,1\}^{4N \times 2}$, where 0 (or 1) indicates whether it originates from a locality sample. The features of these demonstrations $F \in \mathbb{R}^{4N \times d}$ are exacted by $h$, where $d$ is the dimension of features. These are projected into higher dimension $F_{p} = F W_r \in \mathbb{R}^{4N \times M}$ by a randomly initialized weight $W_r \in \mathbb{R}^{d \times M}$, where $M$ is the projected feature dimension. The projected features $F_{p}$ are used to train a classifier $W^*$. Using the form of least squares problem with penalty term, an appropriate classifier weight $W^*$ can be obtained by solving

\begin{equation}
  \begin{aligned}
    W^* = \underset{W}{arg \, min} \, \lVert Y - F_{p} W \rVert_2^2 + \lambda \lVert W \rVert_2^2,
  \end{aligned}
 \label{eq:linear_classifier}
\end{equation}
The solution to the above problem is
\begin{equation}
  \begin{aligned}
        W^* = (F_{p}^{\top} F_{p} + \lambda I)^{-1} F_{p}^{\top} Y
  \end{aligned}
 \label{eq:linear_classifier_weight}
\end{equation}
where $\lambda$ is a coefficient of penalty term, and $I \in \mathbb{R}^{M \times M}$ is an identity matrix.


To reduce memory usage, we store a subset of training samples in a text memory $M_1$. Additionally, to enhance the classification accuracy of $W^*$, some hard-to-classify external samples' questions are stored as $M_2$.

During Inference, for each test sample $(i, x, y)$, we first determine whether it requires updating by comparing its question $x$ to those in $M_2$, and classify by $W^*$. If the maximum similarity between $x$ and $M_2$ doesn't exceeds a threshold $T$, and it's classified as in-domain data, we retrieve $k_0$ similar demonstrations $\{s_i\}_{i=1}^{k_0}$ from $M_1$. These, combined with a demonstration $s_o$ constructed from $x, y$, form a new question $x_{new}=[s_1; s_2;\cdots; s_{k_0}; s_o; x]$, which is then input as $(i, x_{new})$ to the model to obtain the updated output $f(i, x_{new}; \theta_o)$. Otherwise, we use the original model output $f(i, x; \theta_o)$.


\section{Experiments}
\begin{figure*}[t]
\centering
\includegraphics[width=0.9\textwidth]{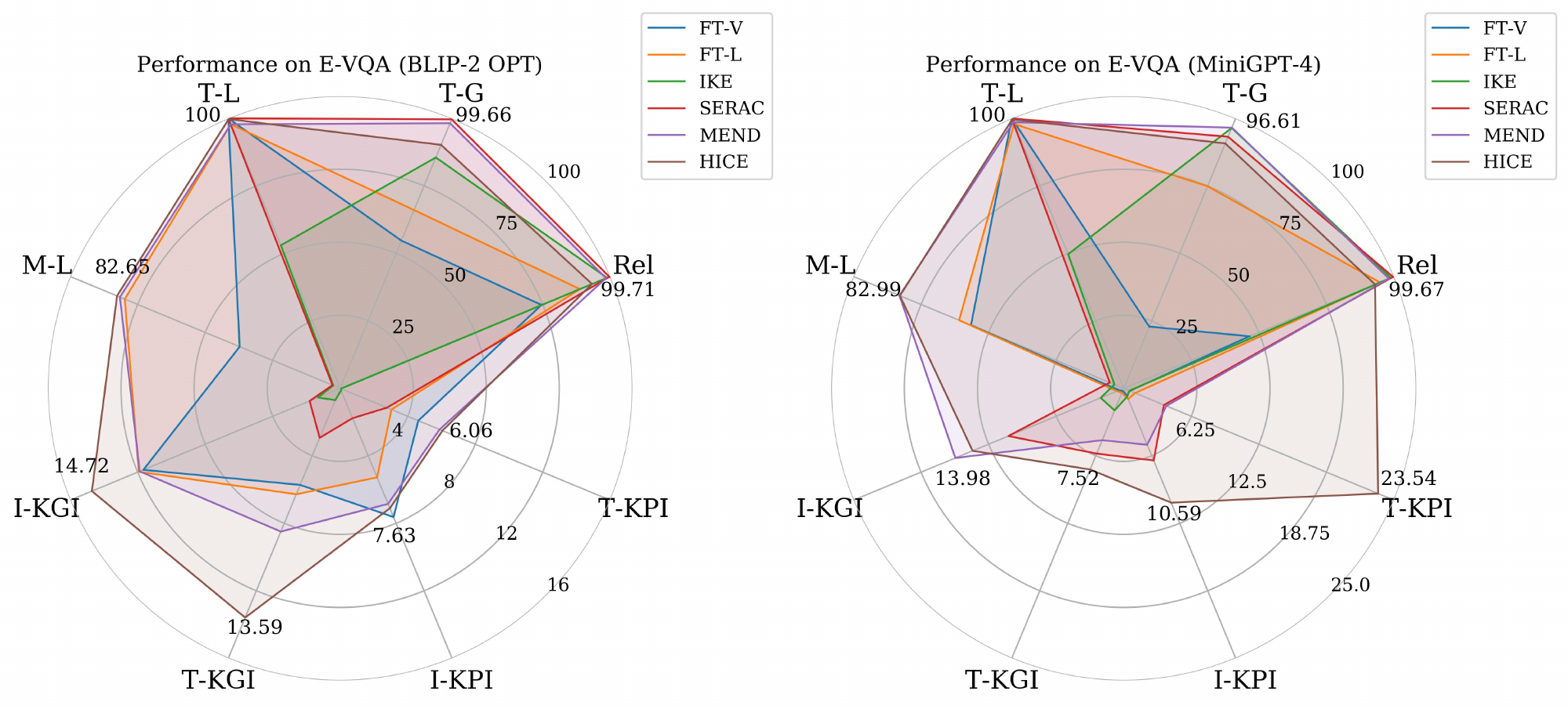}
\caption{Performance comparison of knowledge editing methods on E-VQA benchmark. The range of values for Rel, T-G, T-L, M-L on two backbones are [0, 100], while the ranges of values for I-KGI, T-KGI, I-KPI, T-KPI are [0, 16] on BLIP-2 OPT and [0, 25] on MiniGPT-4.}
\label{fig:vqa_res}
\end{figure*}

\begin{figure*}[t]
\centering
\includegraphics[width=0.9\textwidth]{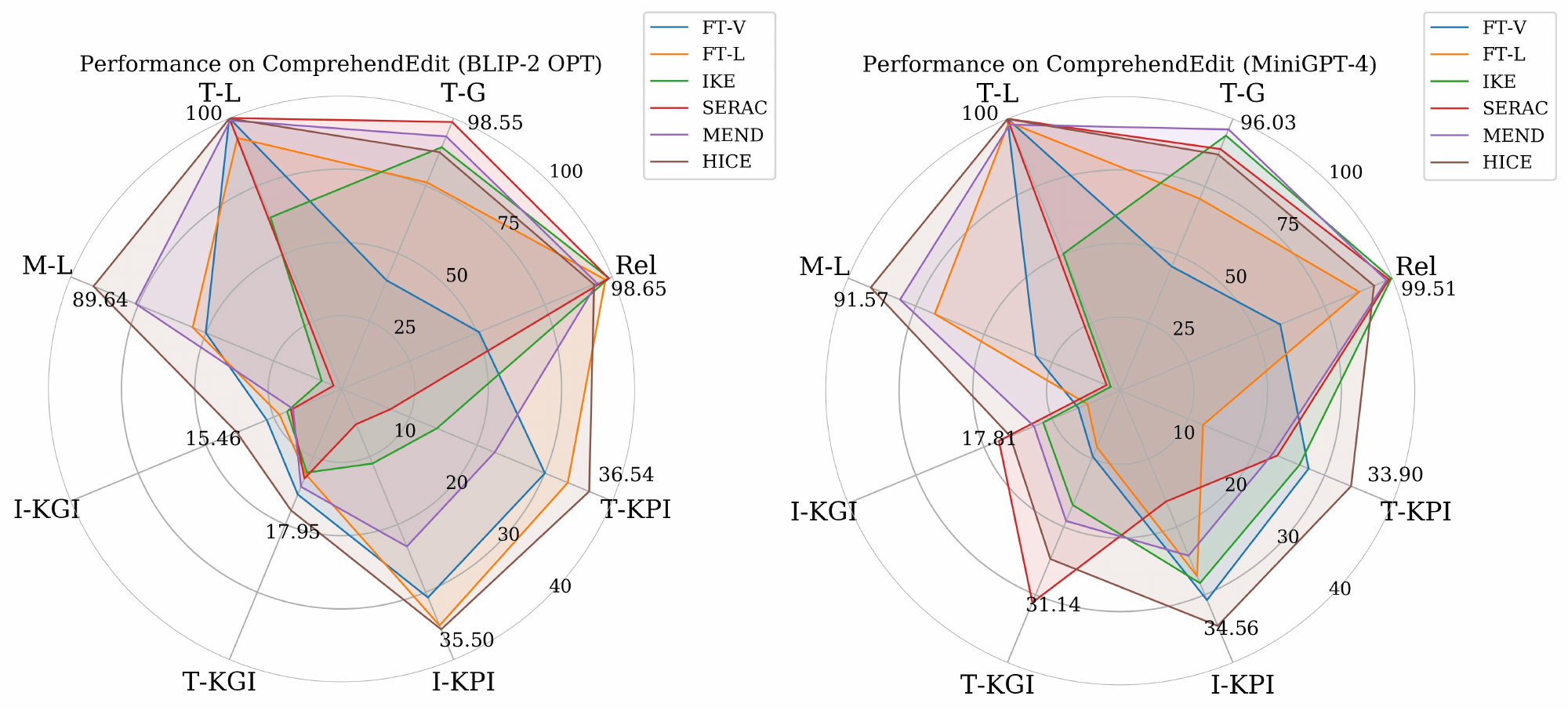}
\caption{Performance comparison of knowledge editing methods on ComprehendEdit benchmark. The range of values for Rel, T-G, T-L, M-L on two backbones is [0, 100], while the range of values for I-KGI, T-KGI, I-KPI, T-KPI is [0, 40].}
\label{fig:our_res}
\end{figure*}

\subsection{Benchmark and Evaluation Metrics}

The evaluation metrics includes \textbf{Rel} (Reliability), \textbf{T-G} (Text Generality), \textbf{T-L} (Text Locality), \textbf{M-L} (Multimodal Locality)~\cite{cheng2023edit} and \textbf{I-KGI}, \textbf{T-KGI}, \textbf{I-KPI}, \textbf{T-KPI}. Due to the content shifts in the rephrased images~\cite{huang2024kebench}, we do not measure multimodal generality.
\subsection{Comparison Methods.}
Our primary comparison targets methods includes Finetune vision model (FT-V), Finetune language model (FT-L), IKE~\cite{zheng2023can}, SERAC~\cite{mitchell2022memory} and MEND~\cite{mitchell2021fast}. The $\dagger$ symbol in the table indicates that we reproduced the results ourselves using the code provided by ~\citeauthor{cheng2023edit} and our own implementation.

\subsection{Implementation Details}
We conduct experiments on PyTorch with NVIDIA RTX 4090 GPUs. For baseline methods, we edit each test sample from the original model by fine-tuning the last layer of the language model (FT-L) and the vision model (FT-V) independently. For other methods like IKE~\cite{zheng2023can}, SERAC~\cite{mitchell2022memory}, MEND~\cite{mitchell2021fast}, we followed the experimental setting described by~\citeauthor{cheng2023edit}. Hyper-parameter values for these methods are provided in the appendix, such as learning rate, optimizer, iteration.

In the process of solving $W^*$, we used $80\%$ of the training samples as training set, and reserved $20\%$ as the validation set. The penalty term's coefficient $\lambda$ is selected from $\{10^{-4}, 10^{-3}, \cdots, 10^{3}, 10^{4}\}$. We choose the one that performed best on the validation set as $W^*$. The dimension of randomly projected features $M$ is set 10,000. The pre-trained language model $h$ is all-MiniLM-L6-v2~\cite{reimers2019sentence}, following~\citeauthor{mitchell2022memory} and the pre-trained CLIP model we use is ViT-B/32~\cite{radford2021learning}. When constructing text memory $M_1$, we use k-means clustering on CLIP-extracted features and select one sample per cluster. The number of cluster is set $5\% \times N$. For each sample we select $k_0=16$ similar samples from memory as context. 

We use BLIP-2 OPT 2.7B and MiniGPT-4 7B to split the original samples into$\mathcal{D}_{KPI}$ and $\mathcal{D}_{KGI}$. When constructing $\mathcal{D}_{KPI}(s)$, $\mathcal{D}_{KGI}(s)$ for each test editing sample $s$, we consider the $k=4$ nearest and farthest neighbors of the test sample $s$, and employ a pre-trained CLIP model~\cite{radford2021learning} to extract features. We use the L2 norm of feature differences as the measure of similarity.

\subsection{Results}
The results of different methods on E-VQA and ComprehendEdit are shown in Fig.~\ref{fig:vqa_res} and Fig.~\ref{fig:our_res}, respectively. These figures reveal that all methods perform significantly on T-L, as the training set differs from out-of-domain data, which indicates that T-L poses little challenge.

ComprehendEdit's data, sourced from multiple datasets, exhibits greater internal variation compared to E-VQA's single-source data. Consequently, the samples in $\mathcal{D}_{KPI}(s)$ differ more from the edited sample $s$, making KPI more challenging in E-VQA.

FT-L and FT-V struggle to perform well on T-G, I-KGI, and T-KGI, as fine-tuning on a single sample limits the model's generalization capabilities. Fig.~\ref{fig:vqa_res} and Fig.~\ref{fig:our_res} demonstrate that indirectly fine-tuning the visual module is less effective than directly fine-tuning the language module, consistent with findings from~\cite{cheng2023edit}.

IKE constructs and selects demonstrations from memory for each test editing sample, combining them as context. This approach performs well on T-G since the context contains similar rephrased questions, which provide effective guidance. However, when processing samples from external datasets, demonstrations constructed from in-domain samples in the context interferes with the output, resulting to inferior performance on T-L and M-L compared to other methods. Moreover, IKE's poor performance on KGI and KPI indicates that the demonstrations used for the editing sample have limited effectiveness on other in-domain samples.

SERAC trains a classifier to decide whether to use the output from the original model or a counterfactual model for a given input sample. It excels on T-L because the questions in the NQ dataset differ significantly from those in the E-VQA dataset, allowing the classifier to identify these external data and rely on the original model's output. However, SERAC underperforms on M-L due to the absence of constraints on multimodal locality during training.

MEND demonstrates strong performance on Rel, T-G, and it especially outperforms other methods on M-L. This is attributed to its use of knowledge distilling loss on external data during the training of an additional module, which preserves existing knowledge after editing. However, its performance on KGI is still limited, since it doesn't use in-domain data when calculating knowledge distilling loss. Additionally, MEND's KPI accuracy of on ComprehendEdit is significantly higher than on E-VQA, because there is a greater difference between the editing sample $s$ and the samples in $\mathcal{D}_{KPI}(s)$ in ComprehendEdit. Consequently, after editing, benefiting from gradient projection, the model's knowledge used to answer questions in $\mathcal{D}_{KPI}(s)$ is less affected.

\begin{table*}[t]
\begin{center}
\caption{The effect of each component of HICE.}
\label{tab:ab_component}
\begin{tabular}{ccccccccc}
\hline
Module & Rel & T-G & T-L & M-L & I-KGI & T-KGI & I-KPI & T-KPI \\
\hline
baseline  & 75.86 & 15.44 & 98.1 & 37.63 & 3.90 & 2.89 & 2.59 & 1.08 \\
+$M_1$  & 95.60 & 91.78 & 97.84 & 36.04 & 13.24 & 7.65 & 8.92 & 46.20 \\
+$M_1$+$W_r$  & 95.94 & 92.26 & 99.64 & 52.56 & 13.29 & 7.65 & 8.85 & 46.20 \\
+$M_1$+$M_2$ & 92.54 & 90.11 & 97.84 & 77.80 & 13.00 & 7.41 & 8.15 & 46.01 \\
HICE & 93.16 & 90.39 & 99.62 & 81.58 & 13.90 & 7.06 & 8.80 & 46.34 \\
\hline
\end{tabular}
\end{center}
\end{table*}

Although previous methods perform well on Rel, T-G, T-L, they are limited in M-L and neglect the edited model in-domain performance. Targeting the poor performannce of existing methods on M-L, KGI and KPI, HICE demonstrates significant advantages across these metrics on various datasets and multimodal language models (MLLMs). HICE achieve a balance between Rel, T-G and T-L, M-L, KGI, KPI. The key to HICE's significant performance on M-L lies in the usage of challenging sample memory $M_2$ and the classifier $W^*$, which accurately determines whether a test sample is related to the edited sample. For unrelated input samples, the model can generate outputs directly, preserving performance on out-of-domain samples. For related input samples, HICE search for similar demonstrations in memory $M_1$ and combine them as context, ensuring correct answers to questions related to the edited sample. 

Despite HICE's advantages in KPI and KGI, there is considerable room for improvement. This limitation primarily stems from the fact that in multimodal models, questions are often closely tied to the input image. Relying solely on text-based context to address similar problems has inherent constraints. HICE shows substantial improvement in KPI because the demonstrations selected based on edited samples have lower correlation with samples further from the domain. As a result, the model can maintain accurate responses to these samples with minimal influence from unrelated demonstrations.

\subsection{Ablation Study}
To validate the effectiveness of each component in HICE, we conducted a series of ablation experiments. These experiments were performed on the E-VQA dataset using MiniGPT-4. For evaluating knowledge generalization index (KGI) and knowledge preservation index (KPI), we selected the 1 nearest and 1 farthest neighbor. The ablation studies of hyperparameters are provided in appendix.

\textbf{The effect of each component of HICE}. As shown in Table~\ref{tab:ab_component}, ``baseline'' means we don't search demonstrations in $M_1$, and don't project features and utilize memory $M_2$. The results of the first and second lines suggest that $M_1$ is the core part of HICE, and the demonstrations constructed from the training set is significantly beneficial for improving most indicators.

The last three lines suggest that $W_r$ and $M_2$ are beneficial to improve the M-L, meaning the model is better at maintaining accurate responses for out-of-domain samples. This improvement occurs because the classifier effectively identifies out-of-domain samples, thus maintaining the model's output on these samples. However, there is a slight decrease in Rel and T-G metrics. This decline is likely due to the misidentification of a small number of in-domain samples, which results in the model not modifying its responses for these in-domain samples. Nevertheless, this slight reduction is acceptable given the substantial improvement in the model's performance on out-of-domain samples.

\section{Conclusion}
This study addresses key challenges in multimodal knowledge editing by introducing ComprehendEdit, a comprehensive benchmark with diverse tasks, and novel metrics - Knowledge Generalization Index and Knowledge Preservation Index - to assess in-domain editing impacts. Our baseline method, Hierarchical In-Context Editing, demonstrates balanced performance across various metrics, revealing unique characteristics of multimodal editing and exposing deficiencies in existing methods. This work provides a robust evaluation framework and baseline, paving the way for more effective editing techniques in large multimodal language models. While significant progress has been made, our study highlights areas for future improvement, particularly in addressing the intricate relationship between questions and images in multimodal contexts, opening new perspectives for advancing the field.

\section{Acknowledgements}
This work is funded by the National Natural Science Foundation of China (62206271, 62076195, 92473112), and the Fundamental Research Funds for the Central Universities (AUGA5710011522), and the Shenzhen Key Technical Projects under Grant JSGG20220831105801004, CJGJZD2022051714160501.


{
    \bibliography{aaai25}
}

%

\newpage
\section{Appendix}

In this section, we present the figures and tables referenced in the paper. These include the sample diversity comparison between existing datasets and ComprehendEdit, flowchart illustrating the proposed method HICE, the dataset construction process, experimental parameter settings, ablation studies on various modules and hyperparameters, and examples demonstrating the edited model's performance on adjacent samples.

\begin{figure*}[!t]
\centering
\includegraphics[width=0.8\textwidth]{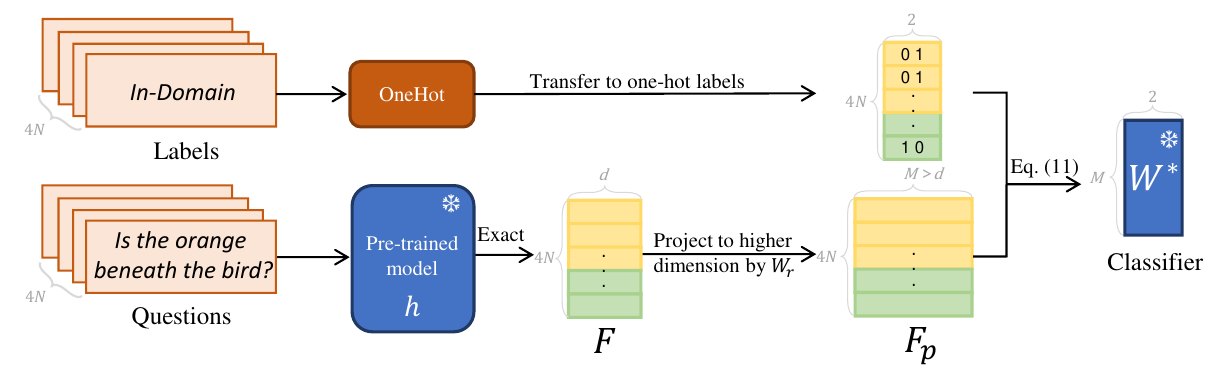}
\caption{Illustration of constructing classifier $W^*$. We first exact features of questions by pre-trained model $h$, and then project these features to obtain $F_p$. $F_p$ are used to calculate $W^*$ by Eq. (11).}

\label{fig:construct_classifier}
\end{figure*}

\begin{figure*}[!t]
\centering
\includegraphics[width=0.95\textwidth]{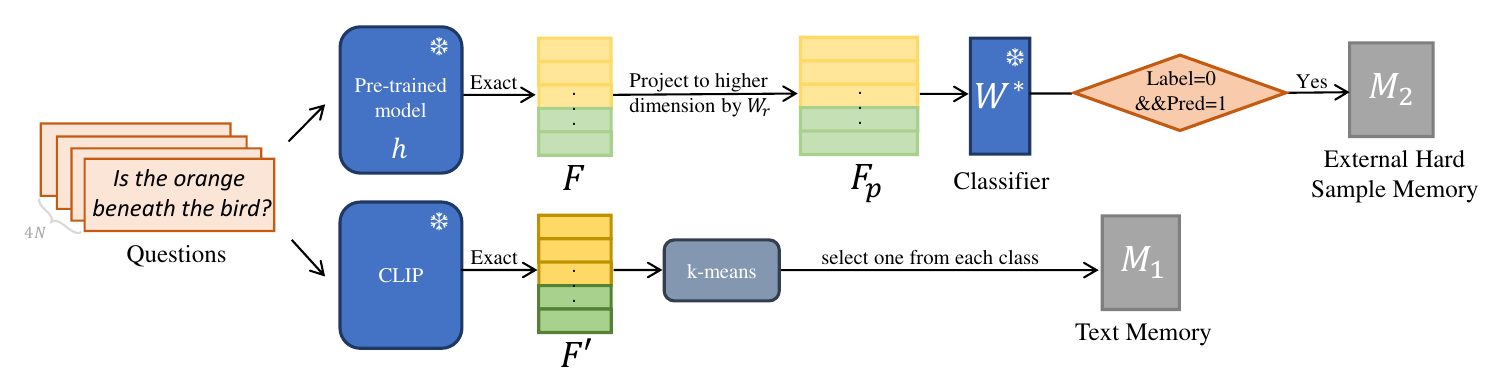}
\caption{Illustration of constructing memories $M_1$ and $M_2$. We use CLIP-extracted features for k-means clustering. Then we randomly select a sample from each class, construct and store the demonstration in $M_1$. We use pre-trained model $h$ and the classifier $W^*$ to make predictions for samples, and then store some hard-to-classify out-of-domain samples in $M_2$.}
\label{fig:construct_memory}
\end{figure*}

\begin{figure*}[!t]
\centering
\includegraphics[width=1\textwidth]{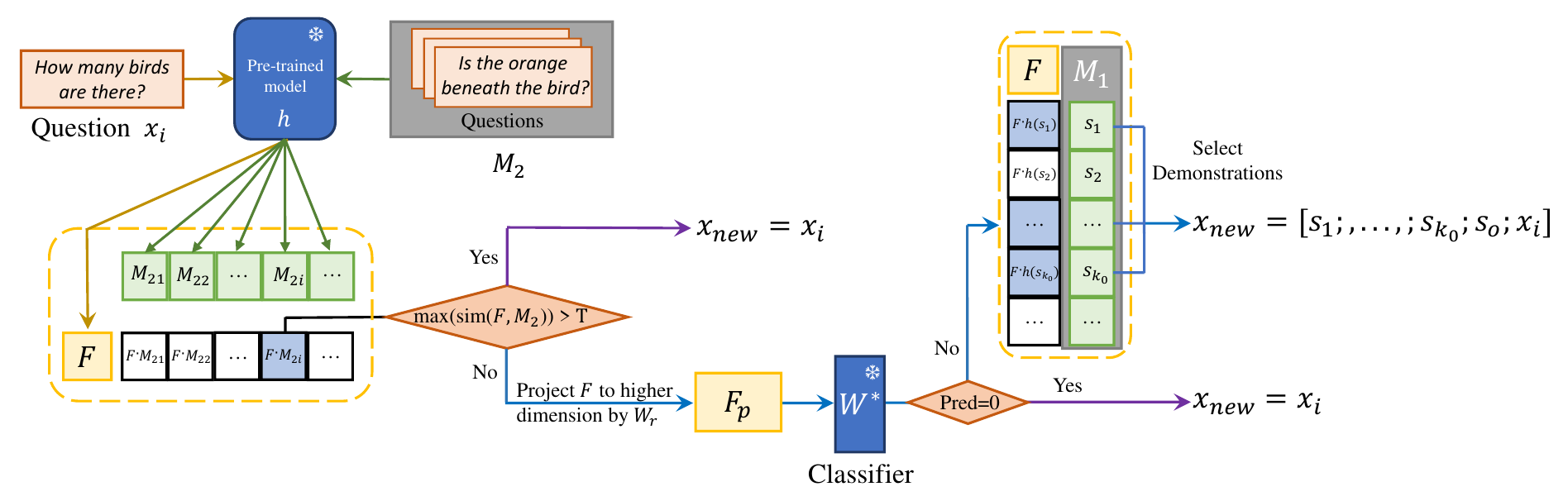}

\caption{Illustration of constructing new question $x_{new}$. For each test question $x_i$, we first determine whether the maximum similarity between it and the sample in $M_2$ is greater than the threshold $T$. If so, and classifier $W^*$ predicts it as an in-domain sample, then $k_0$ demonstrations are selected from $M_1$ to construct the context for $x_i$. Otherwise, the original question will be used $x_{new}=x_i$.}

\label{fig:construct_new_prompt}
\end{figure*}

\subsection{Proposed Method}
We use Llama-2-7b-chat-hf~\cite{touvron2023llama} to generate question types of existing datasets E-VQA and KEBench, the results are shown in Table~\ref{tab:comparison_dataset}. E-VQA and KEBench predominantly focus on object recognition while overlooking other tasks. ComprehendEdit is the first comprehensive multimodal editing benchmark, including 8 tasks derived from diverse datasets.

\begin{table}[h] 
\begin{center}
\caption{Statistics on the number of samples in each task.}
\label{tab:comparison_dataset}
\resizebox{.95\columnwidth}{!}{
\begin{tabular}{cccc}
\hline
Task   & E-VQA & KEBench & ComprehendEdit  \\
\hline
Object Recognition & 4854 & 8089 & 2962 \\
Object Attributes & 1435 & 27 & 2987 \\
Object Counting & 1213 & 0 & 2009 \\
Object Existence & 845 & 3 & 1962 \\
Scene Information & 45 & 44 & 2854 \\
Numerical Inference & 23 & 0 & 846 \\
Spatial Relationship & 16 & 1 & 2239 \\
Text Recognition & 8 & 0 & 2073 \\
\hline
Total & 8439 & 8164 & 17932 \\
\hline
\end{tabular}}
\end{center}
\end{table}

The HICE method mainly consists of three parts: calculating classifiers $W^*$ (Fig.~\ref{fig:construct_classifier}), building memory $M_1, M_2$ (Fig.~\ref{fig:construct_memory}), and constructing corresponding input question $x_{new}$ for each test sample (Fig.~\ref{fig:construct_new_prompt}).

\subsection{Construction of ComprehendEdit}
\begin{figure*}
\centering
\includegraphics[width=0.9\textwidth]{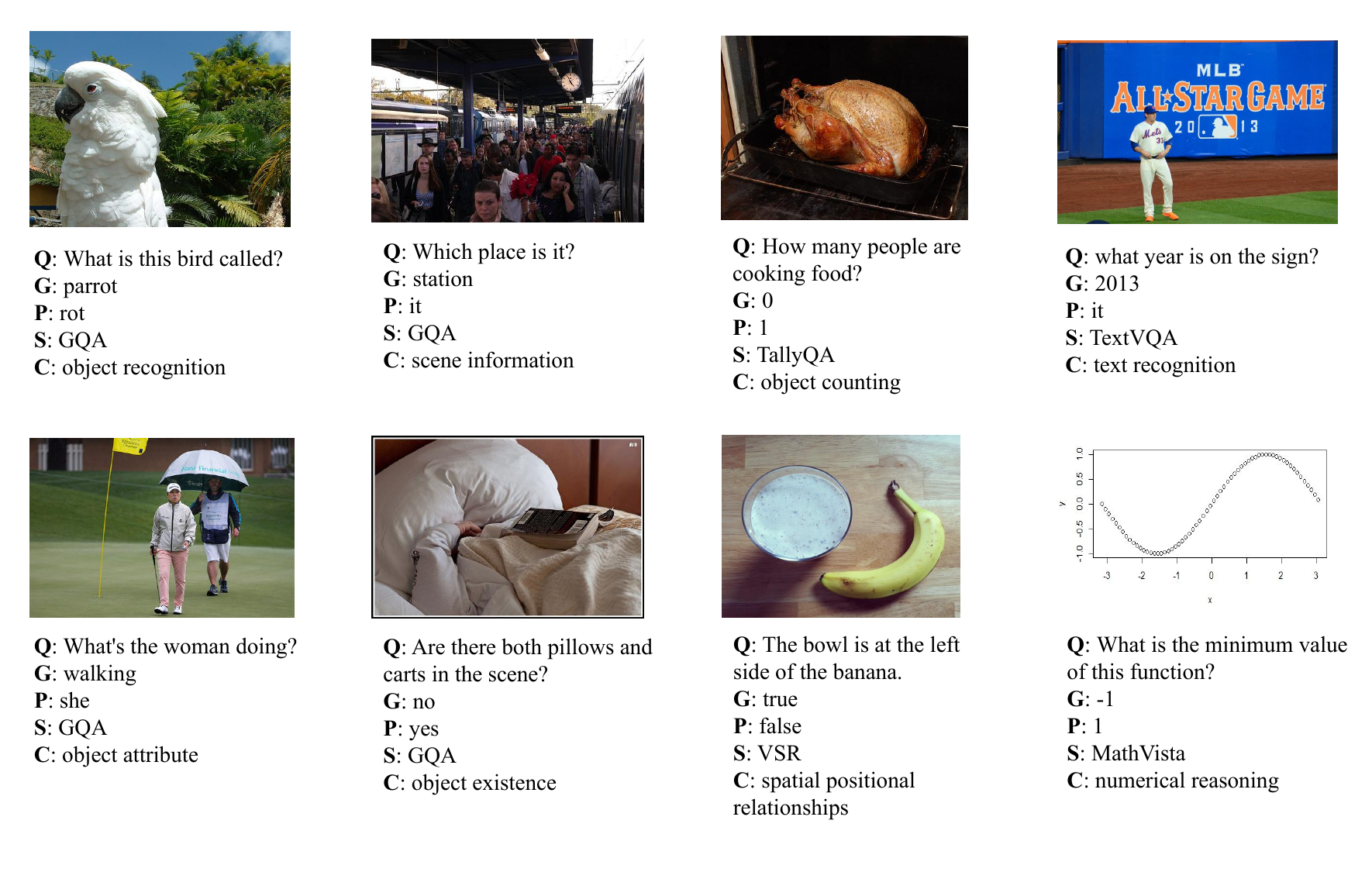}
\caption{Some examples of ComprehendEdit. \textbf{Q}, \textbf{G}, \textbf{P}, \textbf{S}, \textbf{C} mean \textbf{Q}uestion, \textbf{G}round-truth, \textbf{P}rediction, \textbf{S}ource, task \textbf{C}ategory independently.}

\label{fig:our_dataset} 
\end{figure*}

We show some examples of ComprehendEdit in Fig.~\ref{fig:our_dataset}. We selected one sample from each task. Next we will describe the dataset construction process. We used the BLIP-2 OPT 2.7B model and MiniGPT-4 7B model for prediction and initially filtered out samples where both models made incorrect predictions. ComprehendEdit contains diverse subtasks, with samples drawn from various datasets. We will introduce these datasets and how the subtasks were constructed. 

\textbf{GQA dataset}: The GQA dataset comprises 113K images and 22M questions, with images sourced from the COCO and Flickr datasets. Unlike the VQA dataset, GQA addresses biases inherent in VQA, such as the tendency to correctly answer questions based on statistical biases rather than image content (e.g., assuming most tables are wooden). Additionally, GQA evaluates various aspects of model performance, including object and attribute recognition, spatial positional relationships, and scene information.

The validation dataset of GQA are used to construct for four tasks: object existence, object recognition, object attributes, and scene information. The dataset encompasses diverse answers to various questions, with significant variations in answer frequencies. Therefore, when constructing the training and testing sets, both the answer type and the frequency of occurrence were considered.

For instance, in the object recognition task, which comprises 2,962 samples with 359 different answer types, the samples corresponding to the top 127 most frequent answers are selected for the training set. The remaining 232 answers and its questions were used to form the test set. The construction process of object attributes and scene information tasks is the same as above. For the object existence task, where questions are answered with ``true'' or ``false'', a 1:1 ratio between positive and negative answers was maintained in both the training and testing sets to avoid the aforementioned biases.

Ultimately, the ratio of answer types between the training and testing sets is approximately 1:2, while the ratio of samples between the training and testing sets is 3:1.

\textbf{TallyQA dataset}: The TallyQA dataset consists of 287K questions and 165K images sourced from the COCO and Visual Genome datasets. Compared to the VQA dataset, TallyQA presents a greater challenge as it includes both simple samples, which can be answered by an object detector, and complex counting samples. These complex samples require not only object detection but also an understanding of object relationships and attributes, demanding a higher level of reasoning ability.

The construction of the object counting task involved utilizing the test set of TallyQA dataset, where the images are sourced exclusively from Visual Genome. Among the samples where models made incorrect predictions, there are 16 distinct answer types. The distribution of these answer types varies significantly, with some having thousands of associated questions and others only a few. In fact, the distribution of questions across different answer types follows an exponential pattern, indicating a wide range of complexity and diversity.

To avoid the answer bias, a selection process was employed to create training and test sets for evaluation. Specifically, 35 samples were chosen from various answer types to form the test set. For the answer types with fewer than 35 questions, all available samples were included in the test set. Additionally, each answer type had 150 questions included in the training set. For those with fewer than 150 questions, all questions not included in the test set were used in the training set.

As a result, the final training set comprises 1,506 samples, with 861 simple questions and 645 complex questions. The test set encompasses 503 samples, consisting of 321 simple problems and 182 complex problems. This distribution ensures a balanced evaluation of both simple and complex object counting challenges.

\textbf{VSR dataset}: The VSR dataset encompasses over 10K questions, 66 distinct spatial relationships, and 6,940 images sourced from the COCO2017 dataset. In contrast to the previous datasets, the VSR dataset provides a more extensive and diverse range of spatial relationships. It includes both training and testing samples, along with instances containing both correct and incorrect answers, making it a comprehensive resource for model evaluation and training.

The validation sets of VSR dataset were used to create datasets focusing on spatial positional relationships, which include 66 different positional relationships. Among these relationships, the 19 types with the most answered wrong samples were selected for the training set, while the remaining types were allocated to the testing set. Consequently, the training set comprises 1,709 samples, while the test set contains 530 samples.

Additionally,  we observe that over $95\%$ of the samples had ``true'' as the answer, leading to an imbalance. To address this, a preprocessing step was implemented. Approximately half of the samples were randomly selected, and a pre-trained CLIP model was used to extract features and calculate similarity between their relationships. The relationships in these samples were replaced with their most similar counterparts, and the answers were changed to false. This process ensured a more balanced distribution between ``true'' and ``false'' answers.

\textbf{TextVQA dataset}: The TextVQA dataset presents a unique challenge by requiring models to recognize text within images to answer questions. It comprises 28,408 images sourced from the Open Images dataset, accompanied by 45,336 questions. Notably, each question has been annotated by 10 people, ensuring robustness and reliability in the dataset. By addressing the issue of inadequate text recognition in previous datasets, TextVQA provides a valuable resource for advancing research in this area.

The validation sets of TextVQA were used to create tasks focused on text recognition. Each sample contains 10 annotations, but some annotations lack confidence. For instance, a sample with four ``finn'' and five ``finnair'' in annotations is considered not confident. To enhance dataset reliability, questions with more than $80\%$ consistency in human-annotated answers were prioritized, with the most common annotation regarded as the true answer. These questions, having more confident answers, were then randomly divided into training and testing sets, maintaining a 3:1 sample size ratio. This approach ensures that the text recognition dataset is robust and balanced, facilitating the effective evaluation of models in recognizing text within visual contexts.

\textbf{MathVista dataset}: This dataset aggregates a total of 6,414 samples from 28 diverse multimodal datasets. Notably, it introduces a novel evaluation criterion by assessing the numerical reasoning capability of models within a visual context. This is the first dataset proposed specifically to evaluate a model's numerical reasoning ability in visual context.

The testmini dataset of MathVista was used to create the numerical reasoning task. Samples with incorrect answers are randomly divided into training and testing sets. Specifically, the training set comprises 634 samples, while the testing set contains 212 samples.

The prompts we used for each dataset are shown in Table~\ref{tab:prompt_each_dataset}.

\begin{table}[h] 
\begin{center}
\caption{Prompts of each dataset.}
\label{tab:prompt_each_dataset}
\begin{tabular}{ll}
\hline
Dataset   & Prompt \\
\hline
GQA & ``Question: \{\} Short answer:''\\
\multirow{2}*{TallyQA}  & ``Question: \{\} Answer with a number. \\
		~ & Short answer:'' \\
\multirow{2}*{VSR}  & ``Question: Is this description true \\
		~ & or false? Description: \{\} Short answer:'' \\
TextVQA & ``Question: \{\} Short answer:'' \\
MathVista & ``\{\} Short answer:'' \\
\hline
\end{tabular}
\end{center}
\end{table}


\subsection{Experiment Setting}
We list the hyper-parameters for each method, as shown in Table~\ref{tab:hyper_ftv},~\ref{tab:hyper_ftl},~\ref{tab:hyper_serac} and~\ref{tab:hyper_mend}. MaxIter means the max training step; Optimizer is the optimizer we used for updating model; LR is the learning rate; and backbones we use the BLIP-2 OPT and MiniGPT-4. 

\begin{table}[!t]
\begin{center}
\caption{FT-V hyper-parameters.}
\label{tab:hyper_ftv}
\begin{tabular}{ccccc}
\hline
Dataset & MaxIter & Optimizer & LR & Backbone \\
\hline
VQA & - & ASGD & 1e-1 & BLIP-2 OPT \\
VQA & - & ASGD & 2e-2 & MiniGPT-4 \\
Our & - & ASGD & 1e-1 & BLIP-2 OPT \\
Our & - & ASGD & 1e-1 & MiniGPT-4 \\
\hline
\end{tabular}
\end{center}
\end{table}

\begin{table}[!t]
\begin{center}
\caption{FT-L hyper-parameters.}
\label{tab:hyper_ftl}
\begin{tabular}{ccccc}
\hline
Dataset & MaxIter & Optimizer & LR & Backbone \\
\hline
VQA & - & ASGD & 1e-2 & BLIP-2 OPT \\
VQA & - & ASGD & 2e-2 & MiniGPT-4 \\
Our & - & ASGD & 2e-2 & BLIP-2 OPT \\
Our & - & ASGD & 1e-2 & MiniGPT-4 \\
\hline
\end{tabular}
\end{center}
\end{table}
When conducting FT-V and FT-L, we don't train the original model and use each test sample to update the original model to obtain the edited model.

\begin{table}[!t]
\begin{center}
\caption{SERAC hyper-parameters.}
\label{tab:hyper_serac}
\begin{tabular}{ccccc}
\hline
Dataset & MaxIter & Optimizer & LR & Backbone \\
\hline
VQA & 50,000 & Adam & 1e-5 & BLIP-2 OPT \\
VQA & 20,000 & Adam & 1e-5 & MiniGPT-4 \\
Our & 50,000 & Adam & 1e-5 & BLIP-2 OPT \\
Our & 20,000 & Adam & 1e-5 & MiniGPT-4 \\
\hline
\end{tabular}
\end{center}
\end{table}

\begin{table}[!t]
\begin{center}
\caption{MEND hyper-parameters.}
\label{tab:hyper_mend}
\begin{tabular}{ccccc}
\hline
Dataset & MaxIter & Optimizer & LR & Backbone \\
\hline
VQA & 30,000 & Adam & 1e-6 & BLIP-2 OPT \\
VQA & 30,000 & Adam & 1e-6 & MiniGPT-4 \\
Our & 30,000 & Adam & 1e-6 & BLIP-2 OPT \\
Our & 30,000 & Adam & 1e-6 & MiniGPT-4 \\
\hline
\end{tabular}
\end{center}
\end{table}

\subsection{Ablation Study}
To assess the HICE's sensitivity to various hyperparameters, we conducted a series of ablation experiments. These experiments were performed on the E-VQA dataset using MiniGPT-4. For evaluating knowledge generalization index (KGI) and knowledge preservation index (KPI), we selected the 1 nearest and 1 farthest neighbor.

\begin{table*}[h]
\begin{center}
\caption{The effect of the size of $M_1$.}
\label{tab:ab_ratio}
\begin{tabular}{ccccccccc}
\hline
Ratio ($\%$) & Rel & T-G & T-L & M-L & I-KGI & T-KGI & I-KPI & T-KPI \\
\hline
1  & 93.16 & 90.15 & 99.64 & 82.80 & 14.20 & 8.44 & 7.48 & 46.80 \\
5  & 93.16 & 90.39 & 99.62 & 81.58 & 13.90 & 7.06 & 8.80 & 46.34 \\
20 & 93.21 & 90.63 & 99.64 & 82.82 & 14.91 & 8.54 & 9.68 & 45.87 \\
\hline
\end{tabular}
\end{center}
\end{table*}

\textbf{The effect of the size of $M_1$}.  Table~\ref{tab:ab_ratio} shows the effect of storing ratio ($\%$) of the training set on various indicators. The size of ratio has little effect on Rel, T-L, and M-L because a demonstration is constructed for each sample to be edited, primarily aimed at correcting the answers of the edited sample. As a result, Rel is not influenced by the Ratio. Meanwhile, The mode of rehprase questions is relatively simple, so only $1\%$ of the training set is sufficient for the model to answer rephrase questions, consistently maintaining a relatively high level of T-G accuracy. Additionally, since the memory $M_1$ only work during testing, ratio does not impact the classifier's performance. So the model generally maintains its answers for out-of-domain samples.


\begin{table*}[h]
\begin{center}
\caption{The effect of threshold $T$.}
\label{tab:ab_threshold}
\begin{tabular}{ccccccccc}
\hline
$T$ & Rel & T-G & T-L & M-L & I-KGI & T-KGI & I-KPI & T-KPI \\
\hline
0.75 & 85.33 & 86.38 & 99.64 & 93.62 & 10.97 & 6.84 & 7.03 & 45.91  \\
0.80 & 92.07 & 90.11 & 99.64 & 84.84 & 12.86 & 7.36 & 7.96 & 46.03  \\
0.85 & 94.93 & 91.97 & 99.64 & 71.10 & 13.19 & 7.53 & 8.58 & 46.15  \\
0.90 & 95.70 & 92.26 & 99.64 & 57.82 & 13.29 & 7.63 & 8.85 & 46.20 \\
\hline
\end{tabular}
\end{center}
\end{table*}

\textbf{The effect of threshold $T$}. Table~\ref{tab:ab_ratio} shows the effect of the threshold $T$. As the threshold $T$ increases, more in-domain data are classified correctly, while more hard-to-classify external samples are incorrectly classified as in-domain. Consequently, Rel, T-G, KGI and KPI increase while M-L decreases as the threshold $T$ rises.

\begin{table*}[h]
\begin{center}
\caption{The effect of projected feature dimension $M$. ``no'' means we don't project the features.}
\label{tab:ab_M}
\begin{tabular}{ccccccccc}
\hline
$M$ & Rel & T-G & T-L & M-L & I-KGI & T-KGI & I-KPI & T-KPI \\
\hline
no    & 92.54 & 90.11 & 97.84 & 77.80 & 13.00 & 7.41 & 8.15 & 46.01 \\
5000  & 92.88 & 90.68 & 99.19 & 81.00 & 13.00 & 7.36 & 8.18 & 46.03 \\
10000 & 93.16 & 90.39 & 99.62 & 81.58 & 13.90 & 7.06 & 8.80 & 46.34 \\
15000 & 92.59 & 90.77 & 99.69 & 94.43 & 13.08 & 7.43 & 8.13 & 46.03 \\
\hline
\end{tabular}
\end{center}
\end{table*}

\textbf{The effect of dimension of projected features $M$}. Table~\ref{tab:ab_M} shows the effect of the projected feature dimension $M$, where ``no'' means random projection was not used. The results show that $M$ significantly impacts M-L. This is primarily because higher feature dimensions make it easier to distinguish out-of-domain data from in-domain data. However, excessively high projected feature dimension can lead to a decrease in Rel.

\begin{table*}[h]
\begin{center}
\caption{The effect of number of selected demonstrations $k_0$.}
\label{tab:ab_k0}
\begin{tabular}{ccccccccc}
\hline
$k_0$ & Rel & T-G & T-L & M-L & I-KGI & T-KGI & I-KPI & T-KPI \\
\hline
4  & 93.07 & 87.28 & 99.62 & 82.88 & 15.04 & 7.51 & 9.40 & 46.73 \\
8  & 93.12 & 89.34 & 99.63 & 82.84 & 14.89 & 6.89 & 9.49 & 46.66 \\
12 & 93.02 & 89.87 & 99.64 & 82.83 & 15.39 & 6.88 & 8.82 & 45.89 \\
16 & 93.16 & 90.39 & 99.62 & 81.58 & 13.90 & 7.06 & 8.80 & 46.34 \\
\hline
\end{tabular}
\end{center}
\end{table*}

\textbf{The effect of number of selected demonstrations $k_0$}. Table~\ref{tab:ab_k0} shows the effect of the number of demonstrations selected from memory. During testing, a demonstration constructed from the editing sample itself to achieve editing, which is not included in $k_0$. This is why the value of $k_0$ has minimal impact on Rel. T-G benefits from larger $k_0$ since more demonstrations contains more valuable examples, which is beneficial for answering rephrased questions. For M-L, if an external data mistakenly classified as in-domain, more demonstrations would be more likely to alter the output. For KGI and KPI, the presence of more demonstrations based on the editing sample has little impact on other in-domain data, since the images or questions differ from the editing samples. Consequently, the demonstrations of the editing samples have limited positive effects on other samples within the domain.

\subsection{Experimental Result}
\begin{figure*}[b]
\centering
\includegraphics[width=1.0\textwidth]{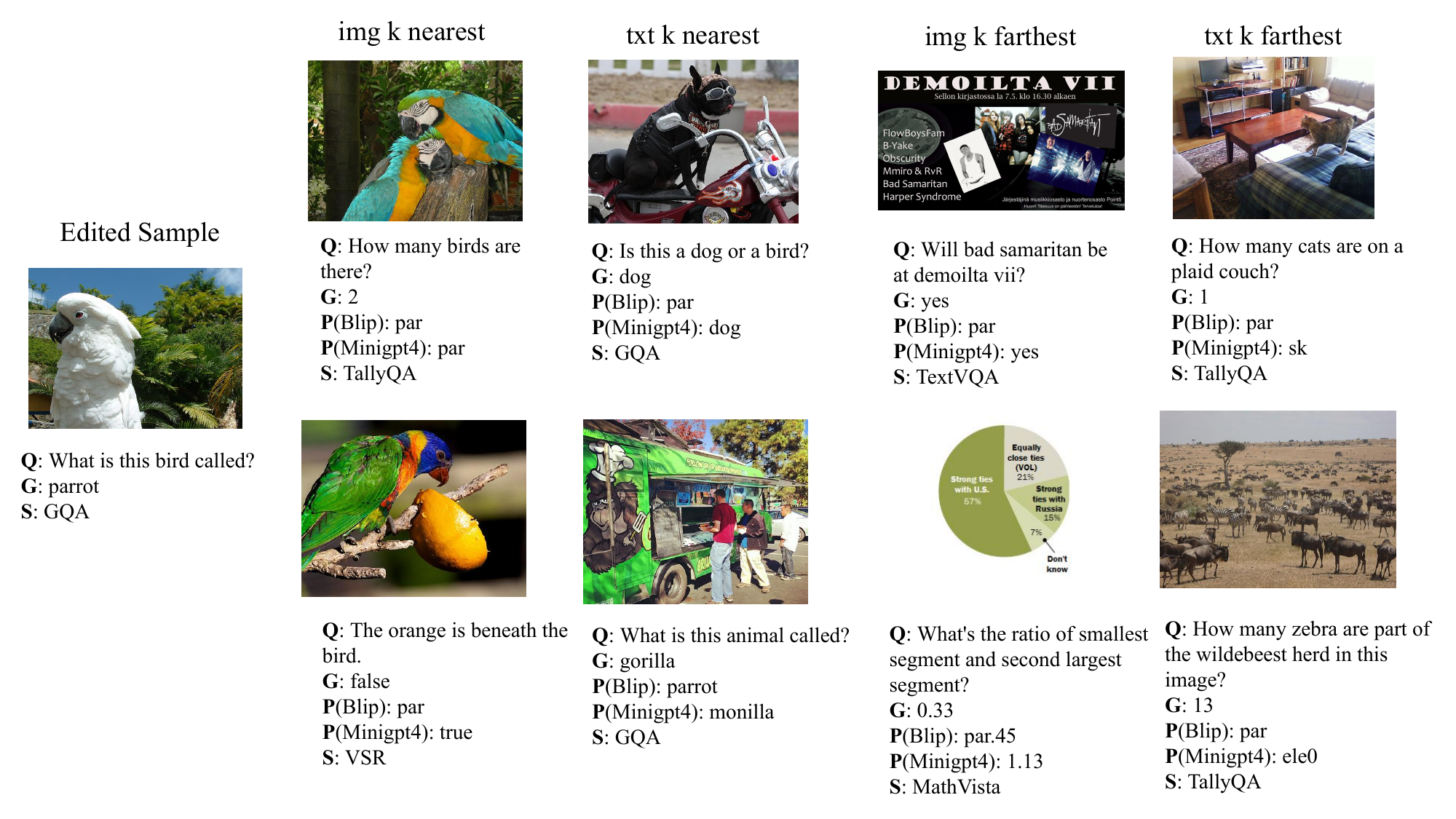}
\caption{Results of the edited model on $\mathcal{D}_{KGI}$ and $\mathcal{D}_{KPI}$ using SERAC. \textbf{Q}, \textbf{G}, \textbf{P}, \textbf{S} mean \textbf{Q}uestion, \textbf{G}round-truth, \textbf{P}rediction, \textbf{S}ource independently. The first row is the performance of edited model on I-KPI and T-KPI, while the second row is the performance of edited model on I-KGI and T-KGI.}

\label{fig:results_after_edit}
\end{figure*}

We show in Fig.~\ref{fig:results_after_edit} the performance of the edited model on KGI and KPI on ComprehendEdit. We present the prediction results of the edited model for some neighboring samples. The first line is KGI performance, and the second line is KPI performance. It can be clearly seen from the figure that the answers of the edited model on other in-domain samples are influenced by the answers of the edited samples.

\end{document}